\documentclass[10pt,twocolumn,letterpaper]{article}

\usepackage{wacv}              

\usepackage{graphicx}
\usepackage{amsmath}
\usepackage{amssymb}
\usepackage{booktabs}
\usepackage{xcolor}
\usepackage{algorithm}
\usepackage{algpseudocode}
\usepackage[accsupp]{axessibility}

\usepackage{mypackages}
\usepackage{mydefs}

\usepackage[pagebackref,breaklinks,colorlinks]{hyperref}

\usepackage[capitalize]{cleveref}
\crefname{section}{Sec.}{Secs.}
\Crefname{section}{Section}{Sections}
\Crefname{table}{Table}{Tables}
\crefname{table}{Tab.}{Tabs.}

\usepackage[pagebackref,breaklinks,colorlinks]{hyperref}

\usepackage[capitalize]{cleveref}
\crefname{section}{Sec.}{Secs.}
\Crefname{section}{Section}{Sections}
\Crefname{table}{Table}{Tables}
\crefname{table}{Tab.}{Tabs.}

\def\wacvPaperID{*****} 
\def\confName{WACV}
\def\confYear{2025}

\begin{document}

\title{Technical Report for ReID-SAM on SkiTB Visual Tracking Challenge 2025}

\author{Kunjun Li, Cheng-Yen Yang, Hsiang-Wei Huang, Jenq-Neng Hwang\\
Information Processing Lab, University of Washington\\
{\tt\small \{kunjun, cycyang, hwhuang, hwang\}~@uw.edu}
}
\maketitle

\begin{abstract}
This report introduces ReID-SAM, a novel model developed for the SkiTB Challenge that addresses the complexities of tracking skier appearance. Our approach integrates the SAMURAI tracker with a person re-identification (Re-ID) module and advanced post-processing techniques to enhance accuracy in challenging skiing scenarios. We employ an OSNet-based Re-ID model to minimize identity switches and utilize YOLOv11 with Kalman filtering or STARK-based object detection for precise equipment tracking. When evaluated on the SkiTB dataset, ReID-SAM achieved a state-of-the-art F1-score of 0.870, surpassing existing methods across alpine, ski jumping, and freestyle skiing disciplines. These results demonstrate significant advancements in skier tracking accuracy and provide valuable insights for computer vision applications in winter sports.
\end{abstract}
\section{Introduction}
\label{sec:intro}
Skiing is a highly dynamic winter sport that demands precision, endurance, and adaptability to rapidly changing environmental conditions. The increasing availability of video footage capturing skier performances in both professional and recreational settings presents opportunities for applying computer vision techniques to advance performance analysis, athlete tracking, and automated feedback systems. However, unlike sports such as soccer or basketball, skiing remains relatively underexplored in the computer vision domain. This is due to challenges such as fast-moving subjects, motion blur, occlusions, and the complex visual characteristics of snow-covered environments.

To address this gap, the SkiTB dataset was introduced as the most extensive and well-annotated resource dedicated to skier tracking and performance analysis. It provides a structured benchmark for evaluating various visual tracking methodologies within the context of skiing. The SkiTB Challenge offers a unique opportunity for researchers to assess the effectiveness of state-of-the-art tracking algorithms and develop novel skier-specific approaches tailored to the complexities of the sport.

In this technical report, we document our participation in the SkiTB Challenge, detailing our methodology for skier appearance tracking. We systematically evaluate various tracking algorithms on the SkiTB dataset and propose optimizations to enhance robustness under the challenging conditions posed by skiing scenarios. This study aims to contribute valuable insights into skier tracking methodologies, emphasizing both algorithmic performance and practical implications for skiing analytics.
\section{Methods}
\label{sec:method}

This submission to the SkiTB Challenge utilizes the recently proposed SAMURAI tracker, enhanced with a re-identification (Re-ID) module and post-processing techniques, to achieve robust skier tracking. The following subsections detail the individual components of our approach: SAMURAI tracker (Section 2.1), the Re-ID module (Section 2.2), and the post-processing techniques (Section 2.3).

\subsection{SAMURAI: A SAM-based Tracker}
We leveraged the state-of-the-art visual object tracking method SAMURAI~\cite{samurai} as our primary tracker. SAMURAI is built upon SAM2~\cite{sam2}, by incorporating a motion-aware memory selection mechanism. This design has demonstrated superior performance on various visual object tracking benchmarks. 

\begin{figure*}[t]
    \centering
    \includegraphics[width=\linewidth]{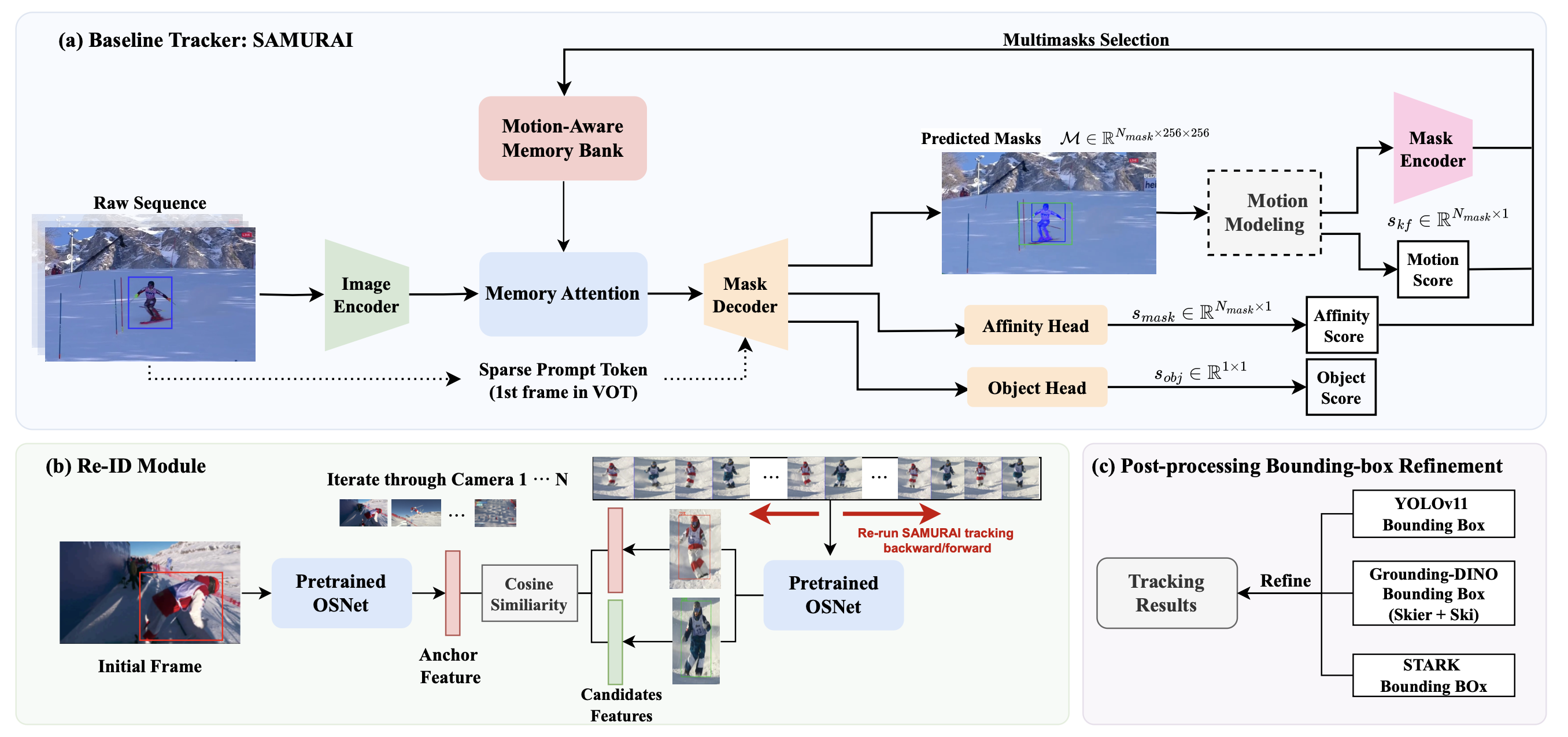}
    \caption{Overall framework of our ReID-SAM.}
    \label{fig:enter-label}
\end{figure*}

\subsection{Re-ID Module}
After obtaining preliminary tracking results from SAMURAI, we performed skier re-identification to correct identity switches that may occur during the tracking process. Firstly, we use a ReID model OSNet~\cite{osnet} to extract the anchor ReID feature $f_{anchor}$ of the target with the given bounding box in the first frame. Secondly, for each camera, we inspect the camera-level ReID feature of the tracking results from SAMURAI and see if they demonstrate high cosine similarity with $f_{anchor}$, if not, we apply a skier detector~\cite{groundingdino} on the middle frame of the camera frames and find out the bounding box $b_{mid}$ that has the most similar ReID feature with $f_{anchor}$. Lastly, based on that bounding box $b_{mid}$, we use it as initial prompt for SAMURAI and conduct forward and backward tracking within that camera to obtain correct camera-level tracking results.

\subsection{Post-Processing}
While the SAMURAI tracker with Re-ID effectively handles skier identification and minimizes ID switching, it primarily focuses on the skier's body, often neglecting equipment like ski boards and poles. To address this, we applied YOLOv11 with a Kalman filter for single-skier scenarios (Alpine (AL) and Jumping (JP)), improving the tracking of both the skier and their equipment by refining bounding box predictions. In multi-skier scenarios, we utilized STARK~\cite{Yan_2021_ICCV} for object detection and calculated the Intersection over Union (IoU) between STARK's predicted boxes and those from the ReID-SAM tracker. If the IoU exceeded a predefined threshold, we adopted the STARK bounding box, ensuring better coverage of both the skier and equipment. Otherwise, we retained the bounding box from ReID-SAM, preserving consistency in the tracking results.
\section{Experimental Results}
\label{sec:exp}

\subsection{Datasets}
The SkiTB dataset~\cite{SkiTBwacv} consists of 300 long videos of professional skiing performances, annotated with dense bounding boxes for visual tracking. It includes videos across three skiing categories: alpine skiing (AL), ski jumping (JP), and freestyle skiing (FS), captured at 161 locations with 196 athletes. The dataset features 352,978 frames in total, distributed across 2019 single-camera clips and multiple-camera setups. Videos are Full HD, and the challenge uses the provided test set, ensuring adherence to the official protocols to evaluate the algorithms' robustness in challenging real-world conditions.

\subsection{Metrics}
Submissions to the challenge are evaluated based on their F1-score, Precision, and Recall. These metrics are reported separately for each of the three categories: AL, JP, and FS. The final F1-score, averaged across these categories, is the key performance indicator.

\subsection{Results}
Table~\ref{tab:result} presents the results of our approach. Our model, ReID-SAM, achieved state-of-the-art performance across all disciplines, surpassing existing methods in terms of F1-score, Precision and Recall. Notably, in the AL and JP categories, our model demonstrated superior tracking accuracy, with significant improvements over competitors such as STARK. Furthermore, our approach exhibited robust performance in the challenging FS scenarios, where the integration of the Re-ID module within ReID-SAM played a crucial role in minimizing identity switches and enhancing long-term tracking stability. These results establish our method as a leading solution for skier tracking on the SkiTB dataset.

\begin{table*}[t]
\fontsize{7}{8}\selectfont
	\centering
	\label{tab:perdisciplinelt}
	\setlength\tabcolsep{.24cm}
	\begin{tabular}{c | c   c  c  c c c c c c c c c c c c | c}
		\toprule
		\rotatebox[origin=c]{90}{Discipline} & \rotatebox[origin=c]{90}{MOSSE} & \rotatebox[origin=]{90}{KCF} & \rotatebox[origin=c]{90}{SiamRPN++} & \rotatebox[origin=c]{90}{FEAR} & \rotatebox[origin=c]{90}{GlobalTrack} & \rotatebox[origin=c]{90}{MixFormer} & 
  \rotatebox[origin=c]{90}{KeepTrack} &
  \rotatebox[origin=c]{90}{OSTrack} & \rotatebox[origin=c]{90}{SeqTrack} &  \rotatebox[origin=c]{90}{LTMU} & 
        \rotatebox[origin=c]{90}{CoCoLoT} & 
  \rotatebox[origin=c]{90}{STARK} & \rotatebox[origin=c]{90}{\yolotr} & \rotatebox[origin=c]{90}{\starkft} & \rotatebox[origin=c]{90}{\algoname} & \rotatebox[origin=c]{90}{\textbf{ReID-SAM}} \\
		\midrule
        \multirow{3}{*}{All} & 0.093 & 0.061 & 0.248 & 0.338 & 0.493 & 0.526 & 0.527 & 0.528  & 0.534  & 0.554  & 0.562 & 0.584  & 0.740 & 0.818 & 0.835 & \bf 0.870 \\
         & \smallresult{0.092} & \smallresult{0.061} & \smallresult{0.270} & \smallresult{0.419} & \smallresult{0.493} & \smallresult{0.518} & \smallresult{0.555} & \smallresult{0.520} & \smallresult{0.538} & \smallresult{0.565} & \smallresult{0.572} & \smallresult{0.595} & \smallresult{0.730} & \smallresult{0.832} & \smallresult{0.843} & \smallresult{0.861}\\
         & \smallresult{0.094} & \smallresult{0.062} & \smallresult{0.235} & \smallresult{0.301} & \smallresult{0.495} & \smallresult{0.535} & \smallresult{0.508} & \smallresult{0.537} & \smallresult{0.533} & \smallresult{0.545} & \smallresult{0.555} & \smallresult{0.576} & \smallresult{0.751} & \smallresult{0.806} & \smallresult{0.829} & \smallresult{0.881}\\

         \midrule

        \multirow{3}{*}{AL} & 0.031 & 0.024 & 0.144 & 0.270 & 0.485 & 0.463 & 0.518 & 0.462 & 0.479  & 0.524 & 0.532 & 0.552 &  0.798 & 0.853 & 0.868 & \bf 0.903\\
         & \smallresult{0.031} & \smallresult{0.024} & \smallresult{0.143} & \smallresult{0.260} & \smallresult{0.487} & \smallresult{0.458} & \smallresult{0.561} & \smallresult{0.457} & \smallresult{0.485} & \smallresult{0.541} & \smallresult{0.546} & \smallresult{0.565} & \smallresult{0.790} & \smallresult{0.874} & \smallresult{0.885} & \smallresult{0.899} \\
         & \smallresult{0.032} & \smallresult{0.024} & \smallresult{0.145} & \smallresult{0.229} & \smallresult{0.483} & \smallresult{0.468} & \smallresult{0.484} & \smallresult{0.467} & \smallresult{0.475} & \smallresult{0.509} & \smallresult{0.521} & \smallresult{0.540} & \smallresult{0.807} & \smallresult{0.834} & \smallresult{0.852} & \smallresult{0.907} \\

        \midrule
        
        \multirow{3}{*}{JP} & 0.155 & 0.098 & 0.281 & 0.373 & 0.504 & 0.574 & 0.536 & 0.577 & 0.590  & 0.576 & 0.584 & 0.603 & 0.818 & 0.880 & 0.896 & \bf 0.919 \\
         & \smallresult{0.153} & \smallresult{0.097} & \smallresult{0.310} & \smallresult{0.451} & \smallresult{0.507} & \smallresult{0.567} & \smallresult{0.576} & \smallresult{0.571} & \smallresult{0.598} & \smallresult{0.591} & \smallresult{0.606} & \smallresult{0.630} & \smallresult{0.807} & \smallresult{0.892} & \smallresult{0.898} & \smallresult{0.915} \\
         & \smallresult{0.157} & \smallresult{0.099} & \smallresult{0.262} & \smallresult{0.338} & \smallresult{0.502} & \smallresult{0.581} & \smallresult{0.510} & \smallresult{0.584} & \smallresult{0.584} & \smallresult{0.565} & \smallresult{0.569} & \smallresult{0.582} & \smallresult{0.830} & \smallresult{0.871} & \smallresult{0.896} & \smallresult{0.923} \\

        \midrule
        
        \multirow{3}{*}{FS} & 0.092 & 0.065 & 0.319 & 0.372 & 0.491 & 0.541  & 0.528 & 0.545 & 0.533  & 0.562 & 0.570 & 0.596 & 0.603 & 0.721 & 0.742 & \bf 0.789 \\
         & \smallresult{0.090} & \smallresult{0.067} & \smallresult{0.358} & \smallresult{0.446} & \smallresult{0.483} & \smallresult{0.529} & \smallresult{0.528} & \smallresult{0.532} & \smallresult{0.530} & \smallresult{0.564} & \smallresult{0.564} & \smallresult{0.590} & \smallresult{0.592} & \smallresult{0.730} & \smallresult{0.746} & \smallresult{0.770} \\
         & \smallresult{0.094} & \smallresult{0.080} & \smallresult{0.298} & \smallresult{0.336} & \smallresult{0.500} & \smallresult{0.556} & \smallresult{0.530} & \smallresult{0.560} & \smallresult{0.539} & \smallresult{0.562} & \smallresult{0.577} & \smallresult{0.604} & \smallresult{0.616} & \smallresult{0.713} & \smallresult{0.738} & \smallresult{0.812} \\

		\bottomrule		
\end{tabular}
	\caption{\textbf{Overall and per-discipline results in the multi-camera (MC) tracking setting.} The \fscore, \prec, and \recall\ scores are presented for each studied algorithm. Our method, ReID-SAM, achieves the best performance across all disciplines.}
    \label{tab:result}
\end{table*}

\section{Conclusion}
ReID-SAM, developed for the SkiTB Challenge, achieves state-of-the-art skier tracking performance by integrating SAMURAI, a Re-ID module, and advanced post-processing. This approach effectively addresses identity switches and equipment tracking challenges, resulting in leading F1-score, Precision, and Recall on the SkiTB dataset. ReID-SAM sets a new baseline for skier tracking and offers substantial potential for enhancing athlete training and performance analysis in winter sports.

{\small
\bibliographystyle{ieee_fullname}
\bibliography{egbib}
}

\end{document}